\documentclass{article} 
\usepackage{iclr2027_conference,times}

\usepackage{amsmath,amsfonts,bm}

\def\eqref#1{equation~\ref{#1}}

\def\1{\bm{1}}

\DeclareMathAlphabet{\mathsfit}{\encodingdefault}{\sfdefault}{m}{sl}
\SetMathAlphabet{\mathsfit}{bold}{\encodingdefault}{\sfdefault}{bx}{n}

\usepackage{graphicx}
\usepackage{wrapfig}
\usepackage{needspace}
\usepackage{booktabs}
\usepackage{multirow}
\usepackage{placeins}
\usepackage{float}
\usepackage{hyperref}
\usepackage{url}

\title{Agentic-Teacher:\\An Agentic Data Engine for Failure-Driven VLA Post-Training}

\author{
{\bfseries
Gongyue Teng\textsuperscript{1},
Yuqing Xie\textsuperscript{1},
Yixian Zhang\textsuperscript{1},
Feng Gao\textsuperscript{2},} \\[0.2em]
{\bfseries
Shu'ang Yu\textsuperscript{2},
Jiaji Rao\textsuperscript{2},
Sukai Wang\textsuperscript{3},
Guanghui Ren\textsuperscript{3},} \\[0.2em]
{\bfseries
Ya Zhang\textsuperscript{1},
Maoqing Yao\textsuperscript{3},
Yu Wang\textsuperscript{1},
Chao Yu\textsuperscript{1,*},
Xinlei Chen\textsuperscript{1,*}} \\[0.6em]
{\normalfont
\textsuperscript{1}Tsinghua University \quad
\textsuperscript{2}Striding AI \quad
\textsuperscript{3}AGIBot}
}

\iclrfinalcopy

\begin{document}

\maketitle

\fancyhead{}
\renewcommand{\headrulewidth}{0pt}

\begin{abstract}

Improving a pretrained vision-language-action (VLA) model requires demonstrations that address its weaknesses and provide effective supervision for learning. Coding agents can autonomously generate demonstrations through environment interaction, yet successful execution alone does not ensure effective supervision for VLA learning. We introduce Agentic-Teacher, an autonomous data engine for failure-driven VLA post-training. Given a VLA checkpoint, a task specification, and configured environment tools, the engine analyzes failed rollouts, selects intervention states, and iteratively develops closed-loop code policies through execution feedback. Validated policies become reusable demonstration generators that collect successful continuations from states where the VLA struggles. To make these demonstrations suitable for learning, Agentic-Teacher introduces temporally correlated action noise to broaden state-action coverage, removes idle frames while preserving gripper transitions, and filters trajectories using the frozen base model’s flow-matching loss. The resulting demonstrations are combined with successful VLA rollouts for post-training, yielding an updated policy that operates without coding-agent assistance. Agentic-Teacher thus connects the weaknesses of a target VLA to both the generation and curation of its training data. In simulation, it improves \(\pi_0\) success rates from 79.4\% to 87.8\% on LIBERO-10, from 80.4\% to 90.6\% on MetaWorld-10, and from 0\% to 30.0\% across three RoboCasa tasks.


\end{abstract}

\section{Introduction}
\label{sec:introduction}

During deployment, vision-language-action (VLA) policies may encounter states outside their training distribution due to environmental noise or accumulated control errors, potentially leading to task failure. DAgger addresses this distribution shift by collecting expert actions at states visited by the learned policy and aggregating them for further training~\citep{ross2011dagger}. Obtaining this supervision from human experts requires repeated monitoring and intervention. Although HG-DAgger and ThriftyDAgger improve takeover mechanisms and intervention allocation, they still depend on human corrections~\citep{kelly2019hgdagger,hoque2022thriftydagger}, making corrective learning costly to scale across tasks and environments.

Advances in multimodal understanding, code generation, and tool use have enabled coding agents to construct and refine robot behaviors through environment interaction. Code as Policies connects language-model-generated programs to perception and control interfaces~\citep{liang2022code}. Within this tool-based execution paradigm, Harness VLA exposes a frozen VLA as a callable manipulation primitive alongside analytic controllers, enabling agents to coordinate learned and programmed behaviors without policy fine-tuning~\citep{zhang2026harness}. To refine behaviors through environmental feedback, CaP-X, Act-Observe-Rewrite, and Agent as Policy incorporate observation, execution, and program revision into iterative feedback loops~\citep{fu2026capx,kumar2026aor,jia2026agentaspolicy}. ASPIRE and Uni-Skill further support skill construction and reuse, enabling subsequent exploration to build on accumulated experience~\citep{lu2026aspire,xie2026uniskill}.

A complementary line of work uses agents to support robot policy learning. RoboClaw coordinates self-resetting rollouts to collect training data and iteratively improve VLA skill policies~\citep{li2026roboclaw}, while ENPIRE enables coding agents to manage real-world policy evaluation and refine training algorithms~\citep{xiao2026enpire}. InSight acquires missing action primitives through guided interaction and incorporates successful demonstrations into VLA training~\citep{wang2026insight}. Recova develops task and recovery behaviors in digital twins and refines separate policies through real-world DAgger. This motivates using coding agents to autonomously construct corrective experts for a target VLA's observed failures, reducing reliance on human-provided corrections. Our focus is on both generating successful corrective demonstrations and adapting them into effective supervision for VLA post-training.
\begin{figure}[H]
\centering
\includegraphics[width=\linewidth]{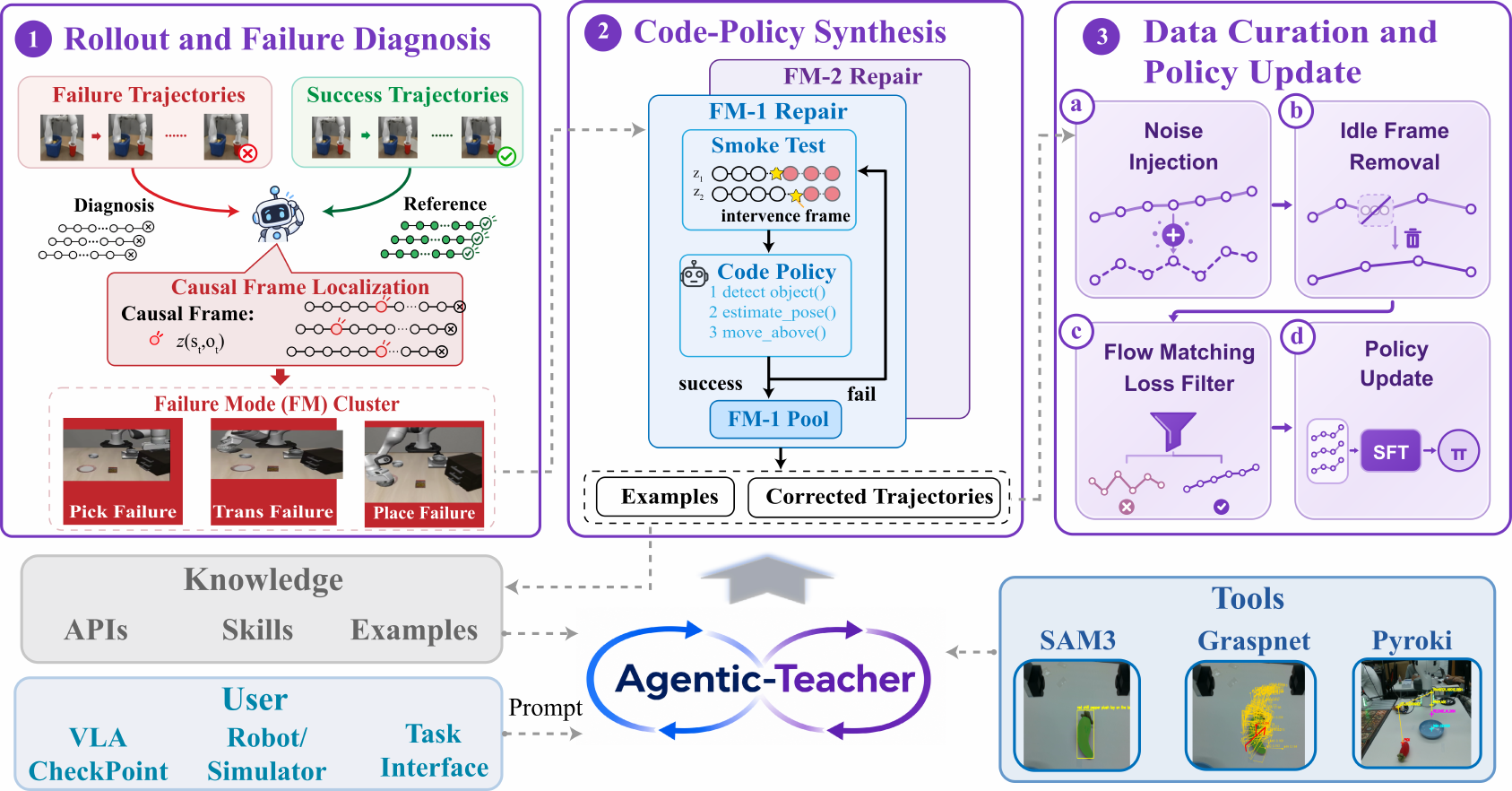}
\caption{Agentic-Teacher pipeline. Given a VLA checkpoint, simulation environment, and task interface, a coding agent uses the harness to evaluate the VLA under varied initial conditions and diagnose failures by comparing successful and failed rollouts. It invokes perception and control tools to synthesize and validate closed-loop code policies that generate corrective demonstrations from VLA failure states. Demonstrations from reliable repair experts are processed through action noise injection, idle-frame removal, and filtering based on the frozen base model's flow-matching loss, and then combined with successful VLA trajectories to construct the VLA fine-tuning dataset.}
\label{fig:method-overview}
\end{figure}

This leads to our central question: \textbf{Can coding agents autonomously identify the causal frame and reason from failure trajectories, generate successful repair demonstrations, and distill the resulting corrective behaviors back into the model?} To answer this question, we introduce \textbf{Agentic-Teacher}, an autonomous data engine that acts as a teacher for failure-driven VLA post-training (Fig.~\ref{fig:method-overview}). Rather than assuming access to a predefined expert capable of correcting every failure, Agentic-Teacher constructs localized code-policy experts on demand, targeting the specific weaknesses revealed by the VLA's own execution.

Given a VLA checkpoint, a task specification, a set of interaction tools, and a success verifier, Agentic-Teacher coordinates failure discovery, corrective policy exploration, data collection, and model evaluation within a shared execution harness. The agents analyze recurring VLA failures, use perception and control tools to synthesize and refine closed-loop code policies, and validate candidate repairs through environment interaction. The resulting successful behaviors provide corrective demonstrations for post-training the VLA. Once updated, the model is evaluated independently without assistance from agents or code policies, while any remaining failures can initiate a subsequent repair cycle.

However, successful code-policy execution does not necessarily yield effective supervision for VLA learning. Program-generated trajectories may exhibit limited behavioral coverage, temporal redundancy, or poor compatibility with the base policy, making task success alone an insufficient criterion for training-data quality. We therefore introduce a trajectory adaptation pipeline that improves behavioral diversity, removes uninformative redundancy, and filters candidate trajectories according to their compatibility with the base model before supervised fine-tuning. In this way, Agentic-Teacher bridges the gap between successful agent execution and learnable corrective supervision, enabling validated repair behaviors to be transferred into the VLA itself.

Our main contributions are as follows:
\begin{itemize}
    \item We introduce Agentic-Teacher, an autonomous data engine that connects VLA weakness identification, local corrective-expert construction, and training-data curation for model post-training. Given a preconfigured environment and tool interfaces, the engine improves the original policy using agent-generated demonstrations without requiring humans to diagnose individual failures, write repair programs, or provide corrective actions.

    \item We develop a trajectory adaptation method tailored to VLA learning. The method addresses the limited behavioral coverage, temporal redundancy, and base-policy incompatibility that can arise in code-policy data by systematically organizing trajectory collection, adaptation, and filtering, thereby making successful program executions more suitable for post-training the student model.

    \item We validate Agentic-Teacher in simulation on LIBERO~\citep{liu2023libero}, MetaWorld~\citep{yu2020metaworld}, and three RoboCasa tasks~\citep{nasiriany2024robocasa}, using 50 evaluation trials per task. Post-training improves the average success rate of $\pi_0$~\citep{black2024pi0} from $79.4\%$ to $87.8\%$ on LIBERO-10, from $80.4\%$ to $90.6\%$ on MetaWorld-10, and from $0\%$ to $30.0\%$ across the three RoboCasa tasks.
\end{itemize}

\section{Related Work}
\label{sec:related-work}

\paragraph{Agents for embodied task solving.}
Embodied agents use different interfaces to connect reasoning with robot execution:
    Code as Policies composes perception and control APIs, while VoxPoser and ReKep
    express motion through spatial value maps and keypoint
    constraints~\citep{liang2022code,huang2023voxposer,huang2024rekep}.
    Agentic-Teacher shares core tools with CaP-X~\citep{fu2026capx}: SAM3 for
    language-conditioned segmentation, Contact-GraspNet for grasp proposal generation,
    and PyRoki for inverse kinematics and motion planning. These tools ground
    agent-generated programs in visual observations and executable robot motions.
    Beyond tool access, ASPIRE and Uni-Skill accumulate reusable skills from
    interaction~\citep{lu2026aspire,xie2026uniskill}. Like ASPIRE, Agentic-Teacher maintains
    a growing skill library: validated repairs contribute reusable code and strategies
    that guide the repair of similar failures.

\paragraph{Autonomous policy repair and improvement.}
Existing approaches differ in the prior experience used to construct corrective behavior.
    InSight first learns a primitive-steerable VLA from existing demonstrations before
    acquiring new primitives through guided interaction~\citep{wang2026insight}.
    UniIntervene learns value estimates and recovery actions from offline trajectories
    and retrieves recovery targets from past interventions~\citep{deng2026uniintervene}.
    Recova combines programmatic and learned recovery, requesting human demonstrations
    when autonomous recovery fails during real-world deployment~\citep{liu2026recova}.
    In Agentic-Teacher, the coding agent constructs repairs by invoking tools and
    consulting prebuilt skill and example libraries. It can therefore develop corrective
    behavior without successful target-task demonstrations or prior execution experience
    on the target task.

\paragraph{VLA post-training.}
DAgger and its human-in-the-loop variants obtain supervision at learner-visited
    states from an external expert~\citep{ross2011dagger,kelly2019hgdagger,hoque2022thriftydagger}.
    Agentic-Teacher automates failure diagnosis and corrective-program construction,
    eliminating per-repair human monitoring and teleoperation after environment and
    tool setup, and thereby substantially reducing the human effort required for
    corrective data collection. RL-based VLA post-training, supported by frameworks
    such as RLinf and RLinf-VLA, can likewise use environment rewards without human
    corrective actions~\citep{yu2026rlinf,zang2025rlinfvla}. Under sparse rewards,
    however, discovering successful behavior is difficult when the initial policy
    rarely succeeds. Agentic-Teacher uses semantic reasoning, reusable skills, and
    targeted program revision to guide exploration toward successful corrections.
    This yields higher successful-trajectory collection efficiency than the reported
    PPO baseline in our RoboCasa comparison (Section~\ref{sec:collection-yield}),
    providing corrective supervision even where reward-driven exploration struggles
    to obtain successful samples.

\section{Agentic-Teacher}
\label{sec:method}

Agentic-Teacher acts as an autonomous teacher for a pretrained VLA by generating
    and curating corrective demonstrations from the VLA's own failures. Given a checkpoint and a task description, it collects rollouts, diagnoses failures,
    and groups episodes that require similar corrections. Coding agents then develop and
    test repair programs for these groups. We process the resulting trajectories and use
    them to fine-tune the VLA. Figure~\ref{fig:method-overview} shows the three stages of
    this process.

\subsection{Rollout and Failure Diagnosis}
\label{sec:localization}

As shown in Figure~\ref{fig:method-overview}, VLA rollouts provide failed trajectories
    for diagnosis and successful trajectories, when available, as references. The agent
    first localizes causal frames and then clusters trajectories with similar failure modes.

\paragraph{Causal frame localization.}
The agent examines the task instruction, camera observations, and action traces to
    identify the interaction responsible for each failure. Successful references help
    distinguish the expected behavior from the failed execution. The agent traces the
    observed failure back to its cause and selects a causal frame at or before the
    faulty interaction as the repair starting point. For example, a placement failure
    may originate in an unstable grasp earlier in the trajectory; localizing this frame
    allows the repair to address the grasp before attempting transport and placement.
    The harness reconstructs the corresponding environment state and checks it against
    the recorded rollout, providing a verified starting point for code-policy exploration.

\paragraph{Failure mode cluster.}
The agent groups failed trajectories within each task according to their causal frames,
    failing stages, and underlying causes. Each group forms a failure mode (FM), such as
    a picking, transport, or placement failure, with supporting trajectory evidence and
    associated repair starting points. This organization allows a coding agent to develop
    and test a targeted repair across similar failures. The resulting failure-mode
    clusters are passed to code-policy synthesis, where each cluster guides the search
    for a reusable corrective program. Appendix~\ref{app:diagnosis} describes the
    diagnosis records and reconstruction procedure.

\subsection{Agentic Code-Policy Exploration}
\label{sec:experts}

\paragraph{Developing a repair program.}
For each failure mode, the coding agent reviews the diagnosis and searches for relevant
    skills, example programs, and tool documentation. Available successful trajectories
    help it understand the intended motion and the order of the remaining task stages.
    The agent uses this information to propose a repair strategy and write a code policy.
    During execution, the program reads observations and calls tools for perception and
    control, adapting its actions as the scene changes. It carries out the remaining task
    from the selected starting state, including the stages that follow the diagnosed failure.

\paragraph{Learning from execution feedback.}
The agent first tests its program on a small set of representative failures. The harness
    returns the task outcome together with videos and action traces, allowing the agent
    to compare the observed behavior with its proposed explanation. If the repair fails,
    the agent examines where the new execution diverges from the intended behavior and
    revises the program. If it succeeds, testing expands to additional starting states
    in the failure group. Successful executions are retained, while unresolved cases guide
    the next revision. When repeated edits bring no improvement, the agent revisits the
    task stage, object grounding, or control strategy rather than continuing to adjust
    the same parameters.

\paragraph{Parallel exploration through a shared harness.}
Separate task agents can develop repairs concurrently, while worker pools execute their
    program evaluations in parallel. The harness provides a common interface for restoring
    starting states, obtaining observations, invoking tools, and checking task completion.
    It also associates each rollout with the program and starting state that produced it.
    Agents can therefore compare attempts and continue exploration from recorded results
    while the workers manage environment execution.

\paragraph{Accumulating reusable skills.}
The experience library stores successful programs and reusable code skills together with
    the tasks and failure modes they address. When developing another repair, the agent
    retrieves relevant entries and adapts them to the current scene and remaining task.
    Records of unsuccessful attempts provide additional context about what has already
    been tried.
    Appendix~\ref{app:expert-construction} details the testing schedule and experience records.

Exploration produces a collection of repair programs and their successful trajectories.
    When the selected starting state is the task initialization, the program completes
    the entire task and produces a full demonstration. The same exploration process can
    therefore collect task data from scratch, using the task description and available
    tools when successful VLA references are absent.

\begin{figure}[H]
\centering
\includegraphics[width=0.94\linewidth]{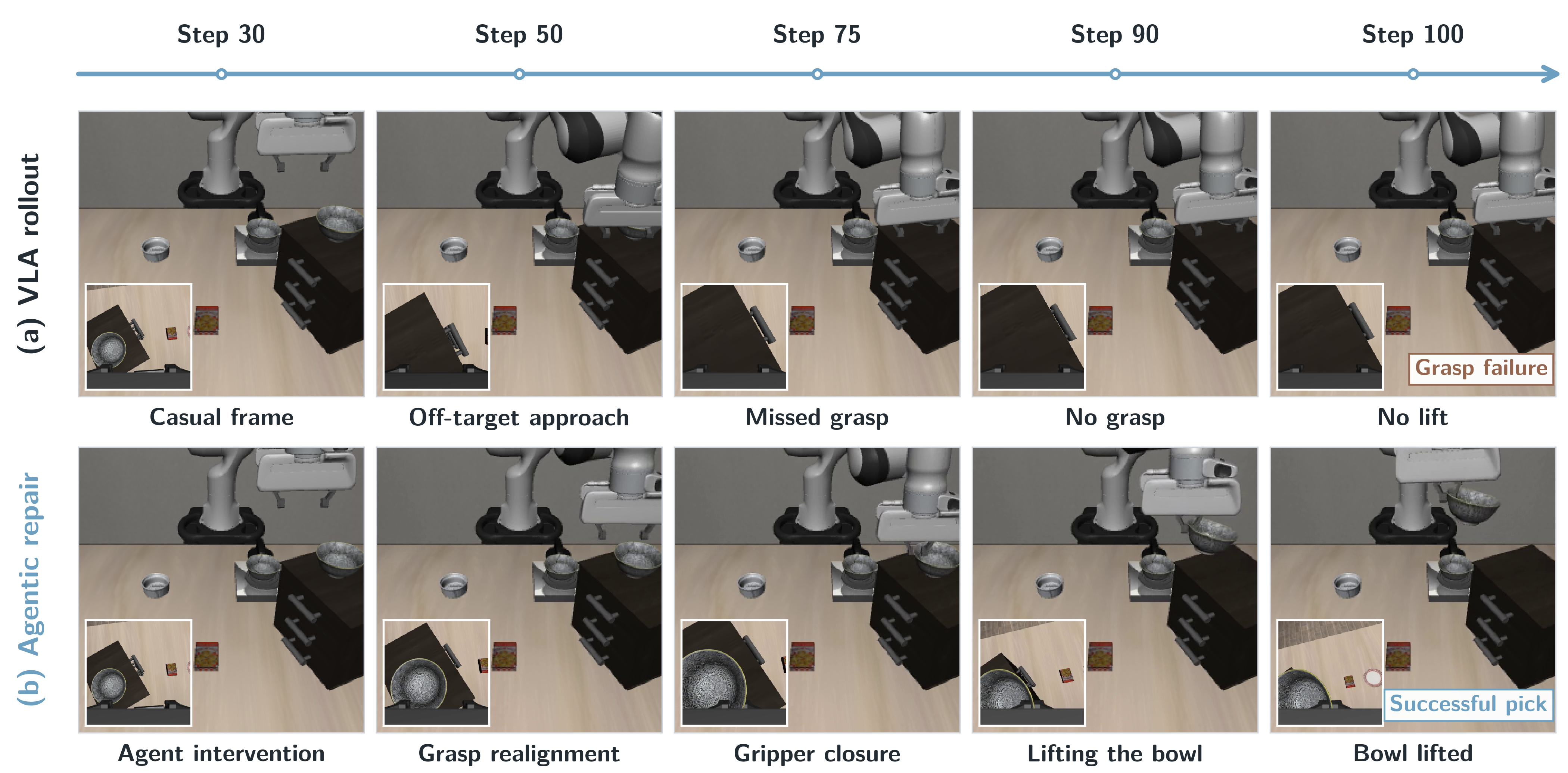}
\caption{A VLA grasp failure and the corresponding code-policy repair.
    (a) The VLA approaches off target, misses the grasp, and fails to lift the bowl.
    (b) The repair program realigns the gripper, grasps the bowl, and lifts it.
    Columns show selected rollout steps, with an additional camera view in each inset.}
\label{fig:pick-repair}
\end{figure}

Figure~\ref{fig:pick-repair} illustrates how a repair can correct an off-target approach
    and recover a successful grasp. The agent identifies the  step 30 as a causal frame and conducts a successful repair demonstration for the pick operation.

\FloatBarrier

\subsection{Data Processing and VLA Post-Training}
\label{sec:corrective-data}
\label{sec:policy-update}

Following the data curation and policy update stage in Figure~\ref{fig:method-overview},
    we convert successful code-policy executions into VLA training data through four
    steps: noise injection, idle frame removal, flow matching loss filter, and
    policy update. These steps broaden behavioral coverage, reduce temporal redundancy,
    and screen trajectories for compatibility with the base policy before fine-tuning.

\paragraph{1. Noise injection.}
We re-execute successful code policies with temporally correlated noise added to the
    arm actions. The perturbations change the states encountered during execution,
    allowing the same program to produce more varied corrective trajectories.
    Gripper commands remain unchanged. Successful rollouts and their corresponding
    observations and actions are retained for dataset assembly.

\paragraph{2. Idle frame removal.}
After joining corrective trajectories to their VLA prefixes, we identify non-idle
    frames from changes between adjacent frames: an absolute end-effector translation
    change greater than 1\,mm along any axis, a relative rotation-vector component
    greater than 3\,mrad in magnitude, or an absolute change greater than 0.1\,mm in
    any finger position.

\paragraph{3. Flow matching loss filter.}
We evaluate each candidate trajectory under the frozen base VLA and exclude trajectories
    with anomalously high flow matching loss. For a trajectory $\tau$, the score is
\begin{equation}
    s_{\mathrm{base}}(\tau)=\frac{1}{K}\sum_{k=1}^{K}
    \mathbb{E}_{u,\epsilon}\!\left[
    \frac{\left\|v_{\theta_{\mathrm{base}}}(A_{k,u},u,c_k)
    -(\epsilon-A_k)\right\|_F^2}{H d_a}\right],
    \label{eq:flow-matching-base-score}
\end{equation}
where $A_k$ is a normalized action chunk of horizon $H$ with $d_a$ physical action
    dimensions, $c_k$ is its observation context, and
    $A_{k,u}=(1-u)A_k+u\epsilon$ interpolates between the actions and Gaussian noise
    $\epsilon\sim\mathcal{N}(0,I)$ at flow time $u$. The frozen model
    $v_{\theta_{\mathrm{base}}}$ predicts the target velocity $\epsilon-A_k$; the score
    averages its squared prediction error over $K$ chunks and sampled noise and times.
    Only physical action dimensions contribute. A high score indicates poor compatibility
    with the base policy's flow matching objective. Appendix~\ref{app:flow-matching-scoring}
    specifies the sampling procedure and estimator.

\paragraph{4. Policy update.}
The retained episodes are aggregated to fine-tune the VLA with its flow matching
    objective. Both retained VLA-prefix and corrective timesteps contribute supervision,
    and successful base-policy rollouts can be included as separate episodes. Action
    chunks respect configured segment boundaries. The updated VLA is evaluated
    independently of the coding agent.

\begin{figure}[H]
\centering
\includegraphics[width=0.94\linewidth,trim=0 4bp 0 0,clip]{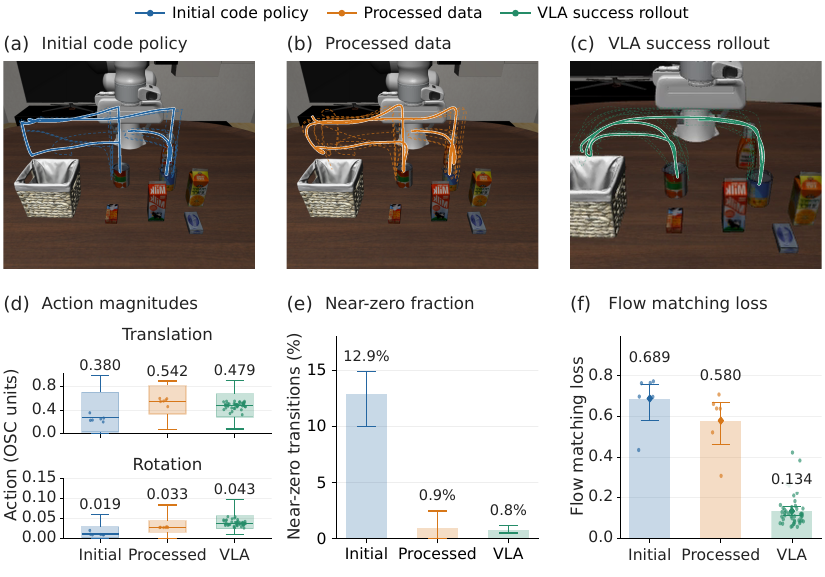}
\caption{Motion and action statistics of initial code-policy trajectories, processed
    training episodes, and successful VLA rollouts. (a--c) Selected end-effector paths.
    (d) Distributions of translation and rotation action magnitudes in OSC units, with
    dots showing episode medians and numbers showing episode-weighted means.
    (e) Fraction of near-zero transitions.
    (f) Flow matching loss; dots denote episode scores and diamonds denote
    group means.}
\label{fig:trajectory-processing}
\end{figure}

Figure~\ref{fig:trajectory-processing} illustrates the differences among initial
    code-policy repair trajectories, processed trajectories, and successful trajectories
    generated by the VLA itself. Panel~(d) compares their action-magnitude distributions:
    the processed trajectories exhibit action magnitudes closer to those of successful
    VLA rollouts. Panel~(e) shows the fraction of idle frames in successful trajectories;
    idle frame removal substantially reduces the proportion of uninformative data in
    the processed trajectories. Panel~(f) further shows that, after noise injection and
    idle frame removal, the processed trajectories have lower flow matching loss
    than the initial code-policy trajectories, suggesting closer alignment with the
    action distribution learned by the base VLA.

\FloatBarrier

\section{Experiments}
\label{sec:experiments}

\subsection{Experimental Setup}
\label{sec:experimental-setup}
We use $\pi_0$ as the base VLA and GPT-5.6 Sol with reasoning effort set to
    \texttt{high} as the coding agent. Experiments cover LIBERO-10, MetaWorld-10, and
    three RoboCasa tasks. For each task, we collect 50 base-policy rollouts and select
    the failed trajectories for agent-guided repair. The repaired
    trajectories are processed and combined with the successful trajectories from
    these initial rollouts to construct the post-training dataset.
    We evaluate the base and updated VLAs
    over 50 rollouts per task using the same evaluation initializations, with seeds
    separate from repair data collection, and report average task success rates for
    each benchmark. Appendix~\ref{app:training-details} reports the initialization checkpoints,
    learning-rate schedules, batch sizes, and training budgets.

\subsection{Downstream VLA Performance}
\label{sec:downstream-performance}
\begin{figure}[H]
\noindent
\begin{minipage}[t]{0.43\textwidth}
\vspace{0pt}
Figure~\ref{fig:benchmark-performance} compares downstream success rates for the
    base checkpoint, Rejected Fine Tune (RFT), and Agentic-Teacher across three
    benchmarks. On LIBERO-10, RFT improves success from
    79.4\% to 81.4\%, while Agentic-Teacher reaches 87.8\%. On MetaWorld-10, RFT
    improves success from 80.4\% to 85.2\%, while Agentic-Teacher reaches 90.6\%.
    Notably, RFT cannot be applied to the RoboCasa tasks because the
    base checkpoint cannot generate successful trajectories for fine-tuning.

\end{minipage}\hfill
\begin{minipage}[t]{0.54\textwidth}
\vspace{0pt}
\vspace{-5.44bp} 
\centering
\includegraphics[width=\linewidth]{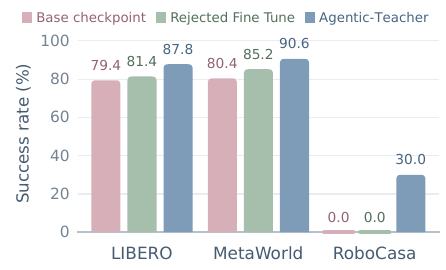}
\caption{VLA success rates across benchmarks.}
\unskip
\label{fig:benchmark-performance}
\end{minipage}
\end{figure}

RFT constructs its fine-tuning dataset exclusively from successful VLA rollouts.
    This reinforces behaviors the VLA already executes successfully, but excludes
    failed rollouts and therefore provides no direct supervision for recovering
    from their failure states. Its ability to expand the policy's coverage of
    out-of-distribution failures is consequently limited. In contrast,
    Agentic-Teacher diagnoses failed executions and synthesizes local repair
    programs to construct corrective
    demonstrations at learner-visited states where the base policy struggles,
    extending the training data beyond its existing successful behaviors.

\paragraph{Failure-mode analysis.}
We conduct a detailed failure-mode analysis on three LIBERO-10 tasks
    (Table~\ref{tab:libero-failure-incidence}). The agent autonomously identifies and
    accurately categorizes failures in the observed trajectories, then develops targeted
    repairs. Most analyzed failure modes are alleviated after post-training. For
    example, handle misalignment in the two-moka-pot task decreases from 22\% to
    0\%, while first-pot approach obstruction and second-pot transport deviation
    each decrease from 12\% to 0\%. Basket-rim collisions in the cream-cheese-and-butter
    task also fall from 4\% to 0\%. These changes suggest that the aggregate gains
    reflect improvements in specific manipulation stages targeted by the repairs,
    rather than only changes in final task outcomes. However, \emph{Butter
    approach obstruction} and \emph{Second-pot approach obstruction} persist near the
    ends of long-horizon tasks, decreasing from 8\% to 2\% and from 10\% to
    8\%, respectively. These residual errors suggest that repair effectiveness
    remains constrained by the base VLA's ability to sustain reliable execution over
    long task sequences.

\begin{table}[!htbp]
\centering
\small
\setlength{\tabcolsep}{3pt}
\renewcommand{\arraystretch}{1.0}
\begin{tabular}{@{}p{0.30\linewidth}lrr@{}}
\toprule
Task & Failure mode & Before repair (\%) $\downarrow$ & After repair (\%) $\downarrow$ \\
\midrule
\multirow[c]{3}{=}{\raggedright Put the cream cheese box and butter in the basket.} & Butter approach obstruction & 8.0 & \textbf{2.0} \\
& Basket-rim collision & 4.0 & \textbf{0.0} \\
& Cheese slippage & 2.0 & \textbf{0.0} \\
\midrule
\multirow[c]{7}{=}{\raggedright Put both moka pots on the stove.} & First-pot setdown rotation & 8.0 & \textbf{0.0} \\
& Second-pot approach obstruction & 10.0 & \textbf{8.0} \\
& Handle misalignment & 22.0 & \textbf{0.0} \\
& First-pot approach obstruction & 12.0 & \textbf{0.0} \\
& Second-pot transport deviation & 12.0 & \textbf{0.0} \\
& First-pot acquisition pressing & 2.0 & \textbf{0.0} \\
& First-pot placement offset & 8.0 & \textbf{0.0} \\
\midrule
\multirow[c]{3}{=}{\raggedright Put the alphabet soup and cream cheese box in the basket.} & Incomplete soup-can release & 2.0 & \textbf{0.0} \\
& Soup-can approach obstruction & 2.0 & \textbf{0.0} \\
& Cheese approach obstruction & 2.0 & \textbf{0.0} \\
\bottomrule
\end{tabular}
\caption{Per-task failure-mode incidence before and after repair on three selected
    LIBERO-10 tasks. All rates use the corresponding task's 50 evaluation trials as
    the denominator.}
\label{tab:libero-failure-incidence}
\end{table}
\FloatBarrier

\Needspace{7\baselineskip}
\subsection{Trajectory Adaptation Ablations}
\label{sec:trajectory-ablation}
\begin{figure}[!htbp]
\centering
\setlength{\abovecaptionskip}{0pt}
\includegraphics[width=\linewidth]{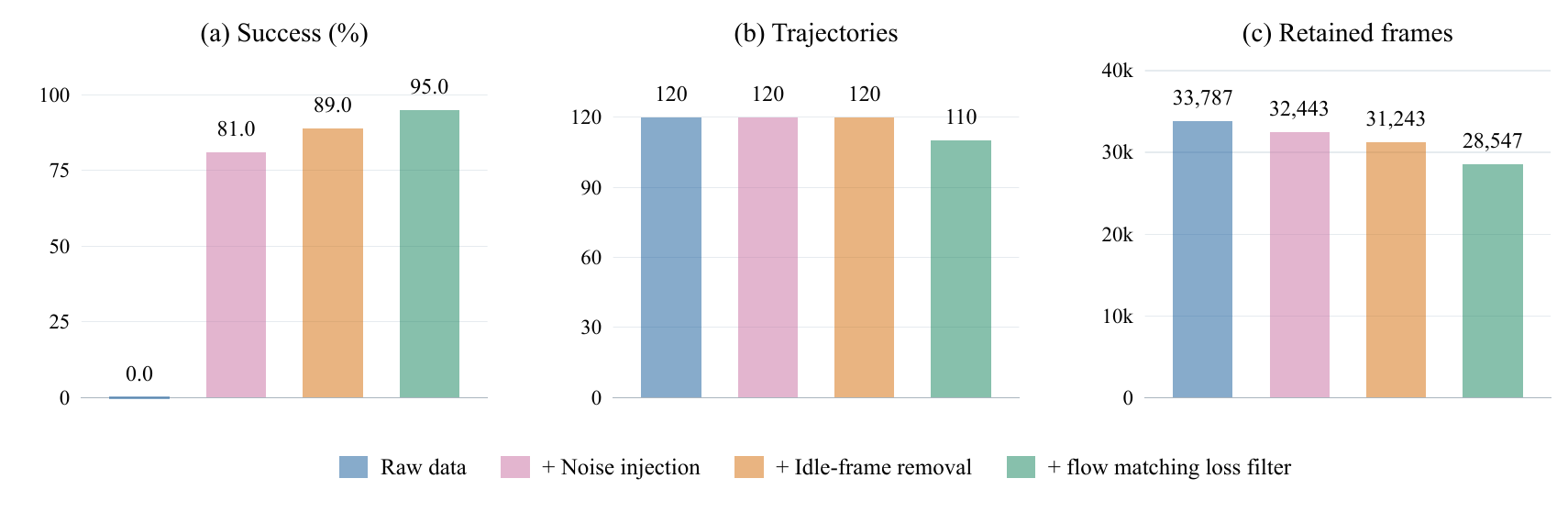}
\caption{Cumulative trajectory-processing ablation on LIBERO at training step 4,000.
    Each successive variant adds the indicated processing step to the preceding variant. Panels show
    (a) task-average success rates, (b) trajectory counts, and (c) retained-frame
    counts.}
\label{fig:trajectory-ablation}
\end{figure}

As shown in Figure~\ref{fig:trajectory-ablation}, training on raw trajectories yields 0.0\% success, whereas the full processing pipeline reaches 95.0\%. This gap likely stems from the highly smooth and repetitive nature of raw code-policy trajectories, which cover only a narrow state-action distribution. During closed-loop rollouts, even small prediction errors can accumulate and push the VLA into out-of-distribution states, where it has little supervision for recovery. These results highlight the importance of adapting agent-generated behavior to provide broader coverage for policy learning.

Across the cumulative variants, success reaches 81.0\% with noise injection,
    increases to 89.0\% after idle frame removal, and reaches 95.0\% with the flow
    matching loss filter. This progression is consistent with the benefits of
    broader behavioral coverage, less redundant supervision, and improved compatibility
    with the base VLA. Notably, the final filtering step improves success while reducing
    the number of trajectories from 120 to 110, indicating that selecting suitable
    demonstrations can improve learning even with less data. Together, these results
    support the effectiveness of the trajectory adaptation pipeline in transferring
    code-policy repairs into the VLA.

\subsection{Successful-Sample Collection Efficiency}
\label{sec:collection-yield}
\begin{figure}[!htbp]
\centering
\includegraphics[width=4.2in]{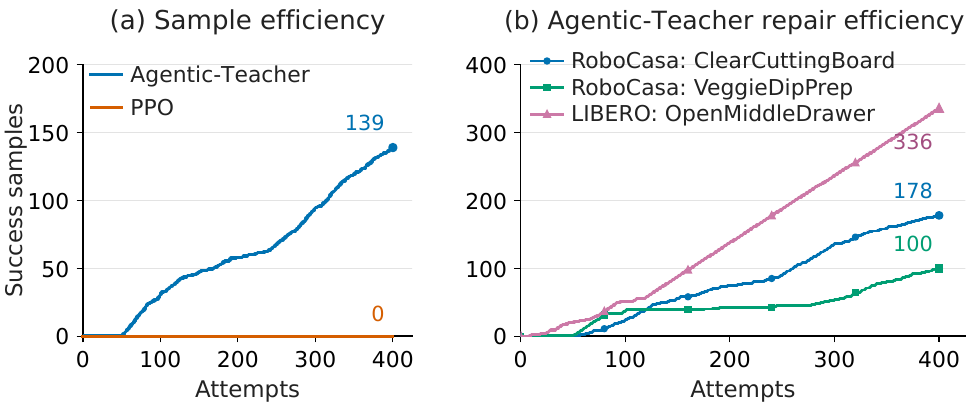}
\caption{(a) Agentic-Teacher versus the
    reported PPO baseline, averaged across RoboCasa tasks. (b) Agentic-Teacher repair
    efficiency on individual RoboCasa and LIBERO tasks.}
\label{fig:sample-efficiency}
\end{figure}

Agentic-Teacher collects successful training candidates more efficiently than the
    reported RL baseline in the evaluated setting (Figure~\ref{fig:sample-efficiency}).
    Within 400 attempts per task, it produces an average of 139 successful trajectories
    across the two RoboCasa collection tasks, while PPO produces none. By constructing
    and refining corrective programs, the framework turns a limited number of attempts
    into successful samples that can be adapted for VLA post-training. This process
    does not require a minimum success rate from the base checkpoint: the coding agent
    can develop corrective behavior even when the initial VLA produces no successful
    target-task rollout.

Exploration efficiency nevertheless depends on the coding agent's capabilities.
    On the complex RoboCasa tasks, Agentic-Teacher produces successful samples at a
    lower rate than in the LIBERO drawer-opening case. When the agent struggles to
    construct reliable corrective programs, more attempts are needed to obtain useful
    supervision.

\subsection{Role of Successful References}
\label{sec:reference-analysis}
\begin{figure}[H]
\noindent
\begin{minipage}[t]{0.51\textwidth}
\vspace{0pt}
Although Agentic-Teacher does not require successful rollouts from the base VLA, successful reference trajectories can improve repair efficiency. In the LIBERO-Goal middle-drawer case, the agent struggles to locate the middle handle accurately and requires extensive exploration to determine suitable gripper poses for grasping and opening the drawer. Successful references provide guidance on both the handle location and the required motion, substantially shortening this exploration process.

In the LIBERO-Object pick-and-place case, the motion sequence is relatively simple, and failures mainly arise from insufficient gripper alignment accuracy during grasping. Successful references therefore offer more limited benefits, as the remaining difficulty lies in precise control rather than identifying the intended behavior. These cases suggest that references are most useful when exploration is hindered by uncertainty about target selection or motion strategy, while fine-grained control errors still require refinement through environment interaction.
\end{minipage}\hfill
\begin{minipage}[t]{0.47\textwidth}
\vspace{0pt}
\centering
\includegraphics[width=\linewidth]{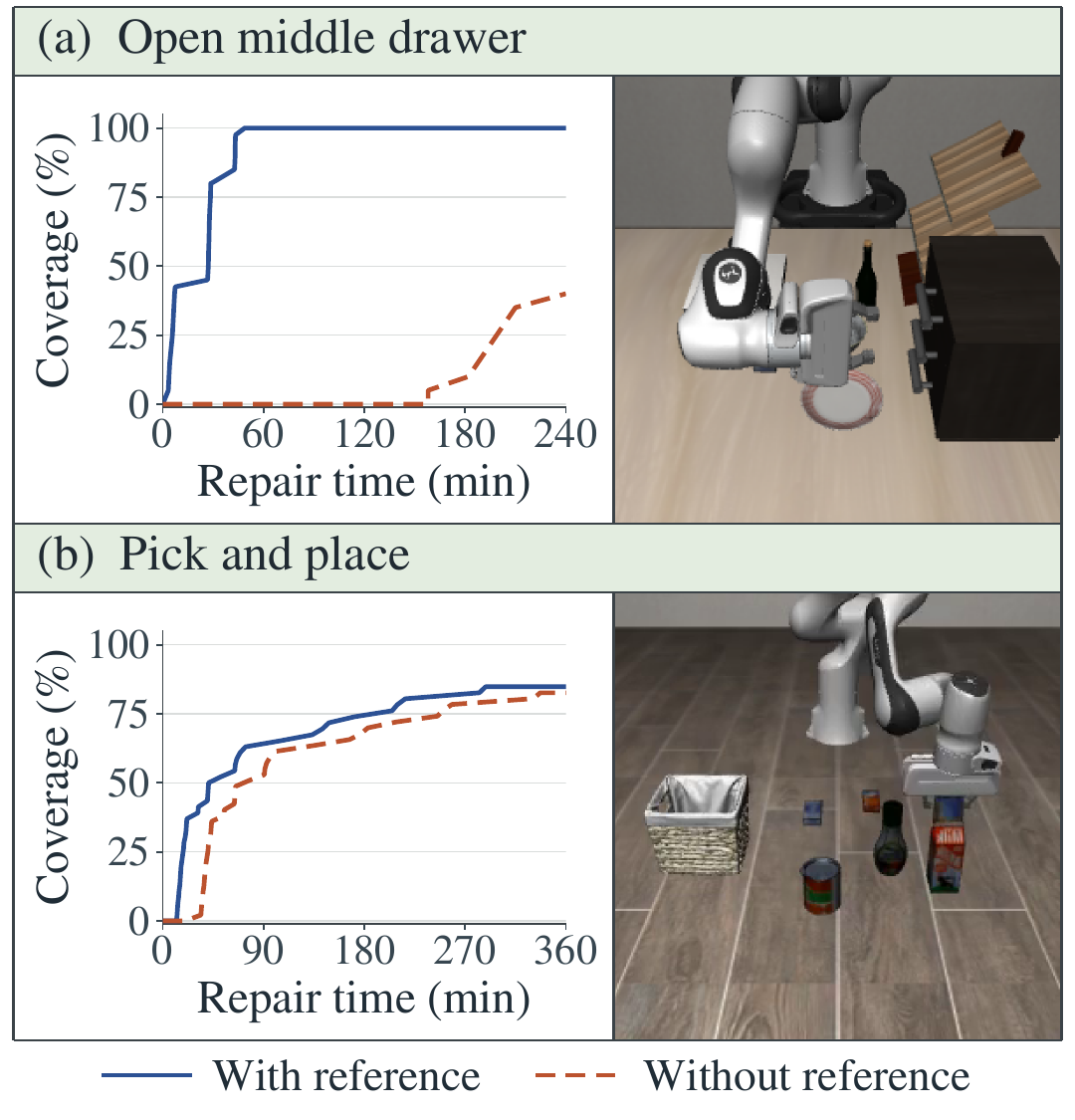}
\caption{Repair coverage with and without successful references: (a) drawer opening;
    (b) pick-and-place.}
\label{fig:reference-efficiency}
\end{minipage}
\end{figure}

\section{Discussion and Limitations}
\label{sec:discussion}
\paragraph{Skill-library coverage and repair-data quality.}
Repair capability depends on the skills available to the coding agent. Tasks involving
    complex end-effector control or fine manipulation can require prolonged exploration
    and remain difficult to repair when the skill library provides insufficient support.
    The quality of repair data also strongly affects post-training: even a successful
    code-policy trajectory may contain undesirable behavior patterns that, when learned
    by the VLA, can degrade its performance. Task success alone therefore does not
    guarantee useful corrective supervision.

\paragraph{Shared training configuration.}
Optimal training hyperparameters can vary across models and tasks. In our joint-training
    settings, data from multiple tasks share a training configuration,
    which may not be optimal for every task. Training separately for each task and
    tuning its hyperparameters may yield further improvements beyond the current
    joint-training setup.

\paragraph{Replay cost on physical robots.}
Each repair iteration currently requires replaying the failed trajectory up to its
    localized causal frame before testing a revised code policy. On physical robots,
    this entails resetting the scene and re-executing the preceding actions, adding
    time and operational cost to repeated repair attempts. Efficiently reproducing
    these repair starting states remains a challenge for real-world deployment.

\section{Conclusion}
\label{sec:conclusion}
Agentic-Teacher constructs corrective programs from VLA failures and converts their
    executions into post-training data. The reported results show improved
    standalone task performance across three benchmark settings. The data engine connects
    failure-driven teacher construction with trajectory adaptation. Completing the
    component ablations and matched cost comparisons will determine which design choices
    account for the gains and at what collection cost.

\renewcommand{\bibfont}{\small}
\setlength{\bibsep}{3pt plus 0.5pt}
\bibliography{paper}
\bibliographystyle{iclr2027_conference}

\clearpage
\appendix

\section{Failure Diagnosis and Intervention Reconstruction}
\label{app:diagnosis}
The diagnosis records the failing stage, the proposed cause, supporting video frames,
    and the selected repair starting point for each episode. Failure-mode records link
    these episodes to a shared description. When successful references are available,
    the agent compares corresponding task stages across successful and failed executions.
    Otherwise, it uses the task instruction and observable transitions in the failed
    rollouts to identify the required stages.

For stage-entry selection, the agent identifies the first frame of the affected stage
    at which the preceding prerequisites remain satisfied. The record includes the
    preceding stage, inspected frames, and observations supporting the selected boundary.
    A window-based configuration instead selects a frame from an annotated interval
    before or around failure onset. The experiment configuration specifies the selection
    rule and the state-restoration procedure.

The harness reconstructs the environment and replays the recorded VLA action prefix to
    the selected frame. It checks the reconstructed state and observations against the
    source rollout using the configured tolerances. The reset configuration specifies
    whether to preserve the complete dynamic state or initialize a quiescent controller
    state. Each accepted starting point retains its episode, frame, and reconstruction
    record. Failed checks return the episode for diagnosis or reset inspection before
    program evaluation.

\section{Expert Construction and Interaction Accounting}
\label{app:expert-construction}
The research interface follows a study--propose--probe--commit cycle. A study retrieves
    relevant successes, recent failures, and previous attempts for a failure mode.
    Each proposal contains a complete program and an explanation of the intended change,
    including its predicted effect, a possible falsifying observation, and supporting
    evidence. The harness executes an immutable copy of the proposed program.

Probing begins with one representative failure, followed by two additional starting
    states and then the remaining targets. Empty batches are omitted. A failed execution
    stops expansion and supplies feedback for the next proposal. Successful executions
    are committed with their programs and trajectory records; unresolved starting states
    remain available for further exploration. Acceptance requires environment-defined
    task success, completion of the remaining task, and execution without a program error.
    Reset and service errors are handled separately so that program revisions are based
    on valid executions.

If two successive proposals overlap in their targets and produce no new successes,
    the agent revisits an earlier level of its explanation. The levels proceed from
    task interpretation and stage selection to object grounding, motion primitives,
    control strategy, and parameter choices. At the task level, the agent gathers
    additional observations before proposing another program.

Each task has a dedicated agent and an assigned worker pool. The experience library
    maintains a current entry for each task and failure mode, containing the repair
    program, reusable skills, and references to successful executions. Previous attempts
    remain in the execution history. Repair coverage counts starting states with verified
    solutions in this collection. Evaluation on new starting states uses a separately
    specified, frozen program or policy. Interaction accounting includes unsuccessful
    attempts and repeated executions as well as trajectories retained for training.

\section{Data Construction and Training Variants}
\label{app:data-variants}

\paragraph{Action noise during collection.}
Let $\xi_t\sim\mathcal{N}(0,\operatorname{diag}(\boldsymbol{\sigma}^2))$ be independent
    noise innovations over the selected arm-action dimensions. With correlation parameter
    $\rho$ and componentwise bounds $c_{\max}$, the perturbation is
\begin{align}
    \eta_0 &= \operatorname{clip}(\xi_0,-c_{\max},c_{\max}),\\
    \eta_t &= \operatorname{clip}\!\left(
        \rho\eta_{t-1}+\sqrt{1-\rho^2}\,\xi_t,-c_{\max},c_{\max}\right),\quad t>0.
\end{align}
The rollout adds $\eta_t$ to the arm command immediately before the environment step,
    clips the resulting command to the action limits, and records the executed action.
    Gripper commands are unchanged. The implemented OSC configuration uses $\rho=0.85$,
    standard deviations of $0.03$ for translation commands and $0.02$ for rotation
    commands, and bounds of three times the corresponding standard deviations.
    These values apply to normalized action coordinates. Each rollout records the noise
    seed, nominal commands, perturbations, and executed commands.

\paragraph{Gripper-preserving frame removal.}
Let $\Delta p_t$ denote the end-effector translation change, $\Delta r_t$ the relative
    rotation vector, $\Delta g_t$ the vector of finger-position changes, and
    $\Delta q_t$ the gripper-command change between adjacent frames associated with
    source index $t$. For a trajectory of $T$ frames, the retained indices are
\begin{equation}
\begin{split}
\mathcal{K}=\{0,T-1\}\ \cup\ \{t:\;&\|\Delta p_t\|_\infty>1\,\mathrm{mm}
    \ \lor\ \|\Delta r_t\|_\infty>3\,\mathrm{mrad}\\
    &\lor\ \|\Delta g_t\|_\infty>0.1\,\mathrm{mm}
    \ \lor\ \Delta q_t\ne 0\}.
\end{split}
\end{equation}
All differences are computed between adjacent frames in the original sequence, and
    rotation is measured using the relative rotation vector. Each threshold is applied
    component-wise to the absolute change. At boundaries, unavailable differences are
    ignored, while the first and last frames are retained unconditionally. Images,
    states, and actions are selected using the same indices $\mathcal{K}$, and the
    retained rows are reindexed for training. Action values are preserved. The output
    manifest records retained source indices and the fraction of removed frames.

\paragraph{Flow matching objective.}
Let $A\in\mathbb{R}^{H\times D}$ be a normalized action chunk after model-specific padding,
    and let $\mathcal{J}$ index its $d_a$ physical action dimensions, including the gripper.
    Given noise $\epsilon\sim\mathcal{N}(0,I)$ and time $u$, the implemented convention is
\begin{equation}
    X_u=(1-u)A+u\epsilon,\qquad U=\epsilon-A.
\end{equation}
The model $v_\theta(X_u,u,c)$ predicts $U$ given observation context $c$. For a timestep
    validity mask $M$, the chunk loss is
\begin{equation}
    \ell_{\mathrm{flow}}(\theta;c,A,u,\epsilon,M)=
    \frac{\displaystyle\sum_{h=0}^{H-1}M_h
    \sum_{j\in\mathcal{J}}
    \left[v_\theta(X_u,u,c)_{h,j}-U_{h,j}\right]^2}
    {\displaystyle d_a\sum_{h=0}^{H-1}M_h}.
    \label{eq:flow-matching-chunk-loss}
\end{equation}
Only valid physical action targets contribute to the reduction. For a minibatch of $B$
    chunks, fine-tuning uses
\begin{equation}
    \widehat{\mathcal{L}}(\theta)=
    \frac{\displaystyle\sum_{n=1}^{B}\sum_{h=0}^{H-1}M_{n,h}
    \sum_{j\in\mathcal{J}}
    \left[v_\theta(X_{u_n},u_n,c_n)_{h,j}-U_{n,h,j}\right]^2}
    {\displaystyle d_a\sum_{n=1}^{B}\sum_{h=0}^{H-1}M_{n,h}}.
    \label{eq:flow-matching-loss}
\end{equation}
The mask handles padded timesteps; chunks without valid targets are removed before
    sampling. All valid timesteps retained after frame removal contribute to training.

\paragraph{Flow matching loss filter.}
\label{app:flow-matching-scoring}
The scorer freezes the base checkpoint and uses deterministic image preprocessing.
    For each successful code-policy trajectory $\tau$, it selects $K=4$ complete chunks
    of horizon $H=50$. Given $N=T-H+1$ valid starting indices, it divides these indices
    into four temporal bins and takes the midpoint of each bin. Trajectories must supply
    four valid starts. This samples different portions of the execution with equal weight.

For each chunk $k$, we draw $M_{\mathrm{MC}}=32$ pairs of flow times and Gaussian noise:
\begin{equation}
    z_{k,m}\sim\operatorname{Beta}(1.5,1),\qquad
    u_{k,m}=0.001+0.999z_{k,m},\qquad
    \epsilon_{k,m}\sim\mathcal{N}(0,I).
\end{equation}
With $M_h=1$ for every timestep in these complete chunks, the trajectory score is
\begin{equation}
    \widehat{s}_{\mathrm{base}}(\tau)=
    \frac{1}{K M_{\mathrm{MC}}}
    \sum_{k=1}^{K}\sum_{m=1}^{M_{\mathrm{MC}}}
    \ell_{\mathrm{flow}}\!\left(
        \theta_{\mathrm{base}};c_k,A_k,u_{k,m},\epsilon_{k,m},\mathbf{1}\right).
    \label{eq:flow-matching-base-mc}
\end{equation}
Thus, each chunk score averages 32 samples, and the trajectory score averages the four
    chunk scores. The implemented seven-dimensional action configuration reduces over
    the six arm coordinates and the gripper coordinate, excluding model padding.
    Random draws use a base seed of 7 and keys derived from source episode and frame
    indices, keeping the score inputs reproducible when trajectories are regrouped.

The collection pipeline scores code-policy executions after dataset assembly and
    idle-frame removal. Local repairs are joined to their VLA prefixes during dataset
    construction, and scores use the retained frames of the corrective segment.
    Executions that start at task initialization are scored over the retained frames of
    the complete code-policy trajectory. Trajectories with anomalously high scores are
    excluded using the selection rule recorded for each experiment.

\section{Experimental Training Details}
\label{app:training-details}
We initialize each benchmark from its corresponding pre-repair $\pi_0$ checkpoint
    and perform supervised post-training on the constructed dataset. Table~\ref{tab:training-hyperparameters}
    summarizes the initialization, learning-rate schedule, batch size, action horizon,
    and optimization budget. Global batch size is measured across all four GPUs.
    The configured cosine-decay steps are reported separately from the number of
    training steps; a run can end before the configured decay schedule is complete.

\begin{table}[!htbp]
\centering
\small
\setlength{\tabcolsep}{4pt}
\renewcommand{\arraystretch}{1.18}
\begin{tabular}{@{}p{0.25\linewidth}p{0.22\linewidth}p{0.22\linewidth}p{0.22\linewidth}@{}}
\toprule
Parameter & MetaWorld MT10 & LIBERO-10 & RoboCasa \\
\midrule
Initialization & RLinf $\pi_0$ MetaWorld SFT & Pre-repair $\pi_0$ LIBERO checkpoint & $\pi_0$ human300 multitask, step 75,000 \\
Peak learning rate & $1\times10^{-5}$ & $5\times10^{-5}$ & $2.5\times10^{-5}$ \\
Global batch size & 32 (8 per GPU) & 32 (8 per GPU) & 64 \\
Action horizon & 5 & 20 & 20 / 50 \\
Training steps & 20,000 & 10,000 & 5,000 / 10,000 \\
Warmup steps & 1,000 & 200 & 1,000 \\
Cosine-decay steps & 30,000 & 10,000 & 100,000 \\
Minimum learning rate & $2.5\times10^{-6}$ & $1\times10^{-6}$ & $2.5\times10^{-6}$ \\
GPUs / parallelism & 4 / FSDP & 4 / DDP & 4 / FSDP \\
Training seed & 42 & 42 & 42 \\
\bottomrule
\end{tabular}
\caption{Supervised post-training configurations. RoboCasa entries list the action
    horizons and training budgets used across its experiment configurations.}
\label{tab:training-hyperparameters}
\end{table}

The action horizons in Table~\ref{tab:training-hyperparameters} apply to policy
    training. The frozen-base trajectory scorer described in
    Appendix~\ref{app:flow-matching-scoring} instead uses its separately configured
    horizon of 50. Similarly, the training seed of 42 is distinct from the scorer's
    seed of 7. The trajectory-processing ablation in
    Figure~\ref{fig:trajectory-ablation} is evaluated at step 4,000, as specified in
    its caption, rather than at the full LIBERO-10 training budget of 10,000 steps.

\makeatletter
\setlength{\@fptop}{0pt}             
\setlength{\@fpsep}{12pt}            
\setlength{\@fpbot}{0pt plus 1fil}   
\makeatother
\section{Complete Results and Failure Cases}
\label{app:full-results}
\subsection{Per-Task VLA Performance}
\label{app:per-task-performance}
Tables~\ref{tab:libero10-per-task}, \ref{tab:metaworld10-per-task}, and~\ref{tab:robocasa-per-task} report the
    task-level success rates before and after post-training.
    Improvements are not uniform: book-caddy and mug-microwave decrease on LIBERO-10,
    while drawer-open-v3 decreases on MetaWorld-10. All task outcomes are retained
    in the benchmark averages.

\begin{table}[!htbp]
\centering
\small
\renewcommand{\arraystretch}{1.12}
\begin{tabular*}{\linewidth}{@{\extracolsep{\fill}}rlrr@{}}
\toprule
ID & Task & Base $\pi_0$ (\%) & Agentic-Teacher (\%) \\
\midrule
0 & soup-tomato & 88 & 94 \\
1 & cheese-butter & 86 & 98 \\
2 & stove-moka & 86 & 96 \\
3 & bowl-drawer & 96 & 98 \\
4 & mugs-plates & 76 & 90 \\
5 & book-caddy & 96 & 88 \\
6 & mug-pudding & 70 & 80 \\
7 & soup-cheese & 94 & 100 \\
8 & two-moka-stove & 24 & 78 \\
9 & mug-microwave & 78 & 56 \\
\midrule
\multicolumn{2}{l}{Average} & 79.4 & 87.8 \\
\bottomrule
\end{tabular*}
\caption{LIBERO-10 task success rates before and after Agentic-Teacher post-training. Each policy is evaluated with 50 rollouts per task. The average gives equal weight to each task.}
\label{tab:libero10-per-task}
\end{table}

\begin{table}[!htbp]
\centering
\small
\renewcommand{\arraystretch}{1.12}
\begin{tabular*}{\linewidth}{@{\extracolsep{\fill}}lrr@{}}
\toprule
Task & Base $\pi_0$ (\%) & Agentic-Teacher (\%) \\
\midrule
reach-v3 & 40 & 84 \\
push-v3 & 58 & 84 \\
pick-place-v3 & 48 & 70 \\
door-open-v3 & 100 & 100 \\
drawer-open-v3 & 100 & 98 \\
drawer-close-v3 & 100 & 100 \\
button-press-topdown-v3 & 100 & 100 \\
peg-insert-side-v3 & 58 & 70 \\
window-open-v3 & 100 & 100 \\
window-close-v3 & 100 & 100 \\
\midrule
Average & 80.4 & 90.6 \\
\bottomrule
\end{tabular*}
\caption{MetaWorld-10 task success rates before and after Agentic-Teacher post-training. Each policy is evaluated with 50 rollouts per task. The average gives equal weight to each task.}
\label{tab:metaworld10-per-task}
\end{table}

\begin{table}[!htbp]
\centering
\small
\renewcommand{\arraystretch}{1.12}
\begin{tabular*}{\linewidth}{@{\extracolsep{\fill}}lrr@{}}
\toprule
Task & Base $\pi_0$ (\%) & Agentic-Teacher (\%) \\
\midrule
CloseToasterOvenDoor & 0 & 48 \\
VeggieDipPrep & 0 & 30 \\
ClearCuttingBoard & 0 & 12 \\
\midrule
Average & 0.0 & 30.0 \\
\bottomrule
\end{tabular*}
\caption{RoboCasa task success rates before and after Agentic-Teacher post-training. Each policy is evaluated with 50 rollouts per task. The average gives equal weight to each task.}
\label{tab:robocasa-per-task}
\end{table}

\FloatBarrier

\end{document}